\documentclass{article}
\usepackage[utf8]{inputenc}

\usepackage{times}
\usepackage{epsfig}
\usepackage{graphicx}
\usepackage{amsmath}
\usepackage{amssymb}
\usepackage{soul}
\usepackage{bm}
\usepackage{gensymb}
\usepackage{subcaption}
\usepackage{booktabs}
\DeclareUnicodeCharacter{2212}{\textminus}% requires a unicode capable editor

\title{Deep Metric Learning for Computer Vision: A Brief Overview}
\author{Deen Dayal Mohan, Bhavin Jawade, Srirangaraj Setlur, Venu Govindaraju \\
University at Buffalo, Buffalo, New York, USA \\
\{dmohan,bhavinja,setlur,govind\}@buffalo.edu}
\usepackage[
backend=biber,
style=numeric,
sorting=ynt
]{biblatex}
\begin{document}

\maketitle

\section{Abstract}
Objective functions that optimize deep neural networks play a vital role in creating an enhanced feature representation of the input data. Although cross-entropy-based loss formulations have been extensively used in a variety of supervised deep-learning applications, these methods tend to be less adequate when there is large intra-class variance and low inter-class variance in input data distribution. Deep Metric Learning seeks to develop methods that aim to measure the similarity between data samples by learning a representation function that maps these data samples into a representative embedding space. It leverages carefully designed sampling strategies and loss functions that aid in optimizing the generation of a discriminative embedding
space even for distributions having low inter-class and high intra-class variances. In this chapter,  we will provide an overview of recent progress in this area and discuss state-of-the-art Deep Metric Learning approaches.

\textbf{Keywords:} Deep Metric Learning, Triplet Loss, Image Retrieval, Face Verification, Person Re-Identification
\vspace{-5mm}

\section{Introduction}
The field of metric learning is currently an active area of research. Traditionally, metric learning had been used as a method to create an optimal distance measure that accounts for the specific properties and distribution of the data points (for example Mahalanobis distance). Subsequently, the methods evolved to focus on creating representations from data that are optimized for given specific distance measures such as euclidean or cosine distance. Following the advent of Deep Learning, these feature representations have been learned end-to-end using complex non-linear transformations. This has led to the primary research in the area of Deep Metric Learning (DML) to be focused on creating loss/objective functions that are used to train deep neural networks. 

One of the key areas that extensively uses Deep Metric Learning approaches is Computer Vision. This is due to the fact that most computer vision applications deal with scenarios where there is a large  variance in visual features of data samples belonging to the same class. Additionally, multiple samples belonging to different classes might have many similarities in visual features. 

Recent advances in Convolution Neural Networks (CNN) have helped in creating good feature representations from images. While using CNNs as feature extractors under a supervised learning setting, often Softmax-based objective functions are used. Although these loss formulations tend to work well in many applications, they tend to suffer when modeling input data that has high inter-class and low intra-class variances. These properties  of data are present in a variety of real-world applications such as Face Recognition \cite{8756519}, Fingerprint Recognition \cite{9648393} \cite{ridgebase}, Image Retrieval \cite{Mohan_2020_CVPR}, Person Re-Identification \cite{advit}, and Cross-Modal Retrieval \cite{wei2020universal}, \cite{wacvnaploss}. In such a scenario, Deep Metric Learning based loss formulations are used to create highly discriminable embedding spaces.  These embedding spaces are designed so as to have a feature representation of   samples belonging to the same class to be  clustered together and  well separated from clusters of other classes in the manifold.

\begin{figure}[h]
\vspace{2.0em}
\includegraphics[width=12cm]{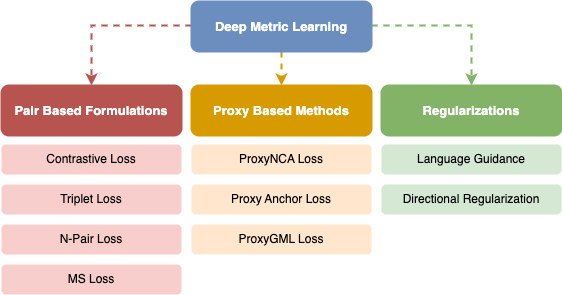}
\caption{An illustration describing various types of deep metric learning losses.}
\label{organization}
\centering
\end{figure}

In this chapter, we discuss popular Deep Metric Learning formulations which are part of the literature. We will restrict our discussion to methods that have found applications in different computer vision tasks. We have organized the different loss formulations into three categories based on the type of overall objective formulations as shown in figure \ref{organization}. The first category consists of pair-based formulations, which formulate the overall objectives based on direct pair-based interactions between samples in the dataset. The second is a group of methods that use a pseudo class representative known as a proxy to formulate the final optimization. Finally, we also discuss regularization methods that try to incorporate auxiliary information that aid in creating more optimal feature representations.

\section{Background}
In this section, we will introduce the mathematical notations and assumptions that are commonly used in deep metric learning literature.  

Consider a dataset $X = \{(x_1,y_1),(x_2,y_2)...(x_n,y_n)\}$, consisting of a set of images and their corresponding class labels. Let $\phi$ be a function parameterized by $\theta$ that maps each image $x_i$ into an embedding space of $d$ dimensions. i.e:
\begin{equation}
\label{neural network}
f_i = \Phi(x_i;\theta_{\phi}); \forall i \in n
\end{equation}
where $f_i \in R^d$, is also referred to as the feature representation of the image $x_i$. Typically, a standard CNN is employed as the feature extractor $\phi$ that produces these feature representations. The overall objective of the feature extractor $\phi$ is to project each image $x_i$ onto a highly separable embedding space, in which all the feature embeddings belonging to a particular class are close to each other and are well separated from the other classes. i.e:
\begin{equation}
\label{eqdistance}
D(f_i,f_j) < D(f_i,f_k); \forall i,j \in y_l; \forall k \notin y_l
\end{equation}
where $D$ is a well defined distance metric in the embedding space. $y_l$ indicates the class label associated with the images. 
For example, if the distance metric under consideration is Euclidean,  (\ref{eqdistance}) can be rewritten as 
\begin{equation}
\label{eucldistance}
 {\|\bm f_i - \bm f_j\|}^2 < {\|\bm f_i - \bm f_k\|}^2 ;  \forall i,j \in y_l; \forall k \notin y_l
\end{equation}
Recently, many feature extractors constrain the final feature representation to have a unit norm. Constraining the feature representation to have unit norms forces the embedding manifold to be an n-dimensional unit hyper-sphere (as shown in the figure). When the feature representation lies on a unit-hypersphere, angular separation is used as the metric to measure the similarity and dissimilarity between the feature representations. Given two feature representations $f_i$ and $f_j$, one could compute the cosine similarity between the two representations. i.e,
\begin{equation}
\label{similarity}
S = \frac{f_i^T f_j}{||f_i||.||f_j||}, i,j \in n \\
\end{equation}
Since the magnitude of the feature representation is 1, the similarity value of the dot product of the representation provides the angular separation between the two feature vectors. The range of $S$ is between -1 and 1, where 1 represents a 0\textdegree angle of separation between the feature representations and -1 represents a 180\textdegree separation. Given a large dataset $X$, the objective \ref{eqdistance} is most often enforced in every mini-batch $B$. Throughout this chapter, we will assume the optimization of deep neural networks using mini-batches.

\pagebreak

\section{Pair-based Formulation}
In this section, we will look at methods that rely on the sampling of informative pairs for better optimization. We will restrict our discussion to a few methods that are widely used in computer vision related applications such as Face Recognition \cite{8756519}, Fingerprint Recognition \cite{9648393} \cite{ridgebase}, Image Retrieval \cite{Mohan_2020_CVPR}, Person Re-Identification \cite{advit}, and Cross-Modal Retrieval \cite{wacvnaploss}.
%etc, as an exhaustive review of these methods, is out of the scope of this chapter. 

\subsection{Contrastive Loss}
As discussed in the previous section, the primary objective of a standard metric learning loss formulation is given by \ref{eqdistance}. One way to achieve this is to enforce the same constraint in the objective function while training a deep neural network. Given two feature representations, $f_i$ and $f_j$ belonging to the same class, the objective is to reduce the distance between the representations. If samples belong to different classes, then the objective is to increase the distance between the feature representations, and this can be mathematically written as: 
\[
    L_{con}= 
\begin{cases}
    {\|\bm f_i - \bm f_j\|}^2 ,& \text{if} \ y_i = y_j\\
    [\alpha - ({\|\bm f_i - \bm f_j\|}^2)]_+,            & \text{else if} \ y_i \neq y_j
\end{cases}
\]
where $y_i$ and $y_j$ are the classes associated with $f_i$ and $f_j$. $\alpha$ here is the margin. We will discuss $\alpha$ in detail in the next section.

\subsection{Triplet Loss}

Triplet Loss, an improvement to the contrastive loss formulation proposed by  \cite{schroff2015facenet}, is a widely used metric learning objective function for creating separable embedding spaces.  Consider three feature representations $f_a$, $f_p$ and $f_n$ corresponding to three images in the dataset. Let $f_a$ and $f_p$ be the feature representations of two images belonging to the same class which we denote as \textit{Anchor} and \textit{Positive} samples respectively. Let $f_n$ belonging to a different class in the dataset be denoted as a \textit{Negative} sample. Triplet loss minimizes the distances between the feature embeddings of the Anchor and the Positive, while maximizing the distance between the Anchor and the Negative. When considering Euclidean distance, the triplet loss is defined as below:

\begin{equation}
\label{eqn_triplet_loss}
\mathcal{L} = \sum_{a,p,n \subset N} \left[ \, {\|\bm f_a - \bm f_p\|}^2 - {\|\bm f_a - \bm f_n\|}^2 + \alpha \, \right]_+
\end{equation}
\begin{figure}[h]
\vspace{2.0em}
\includegraphics[width=12cm]{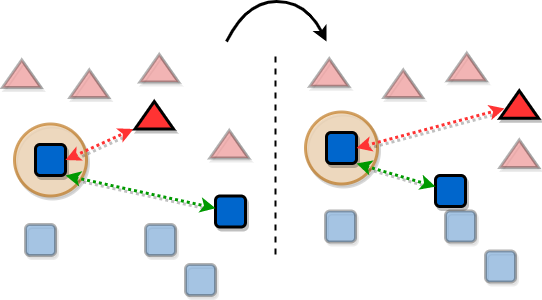}
\caption{Illustration of Triplet Loss. The Positive sample is attracted towards the Anchor  whereas the Negative sample is repelled from the anchor. Here and in all diagrams in this chapter, we represent an anchor with in a brown circle. Red arrow is used to represent repulsion and green arrow is used to represent attraction.}
\centering
\end{figure}

The terms $\bm f_a, \bm f_p, \bm f_n$ correspond to feature embeddings for the anchor, positive and negative samples, where $a, p, n$ are sampled from the training dataset $N$. $\alpha$ defines the margin enforced between the Anchor-Negative distance and the Anchor-Positive distance. $[.]_+$ represents a $max(.,0)$ function.

This mathematical formulation can be thought of as an extension of the Contrastive Loss formulation, as it explicitly enforces Anchor-Positive similarity while repelling the Negative. Additionally, it is also interesting to perform gradient analysis to get a geometric picture of the directions in which these attractive and repulsive forces act. For this we compute the derivatives of the loss (Eq. \ref{eqn_triplet_loss}) with respect to these feature representations as follows:
\begin{equation}
\label{eqn_triplet_loss_grad}
\begin{split}
\frac{\partial \mathcal{L}}{\partial \bm f_a} & = 2 (\bm f_n - \bm f_p) \\
\frac{\partial \mathcal{L}}{\partial \bm f_p} & = 2 (\bm f_p - \bm f_a) \\
\frac{\partial \mathcal{L}}{\partial \bm f_n} & = 2 (\bm f_a - \bm f_n)
\end{split}
\end{equation}

\begin{figure}[h]
\vspace{2.0em}
\includegraphics[width=12cm]{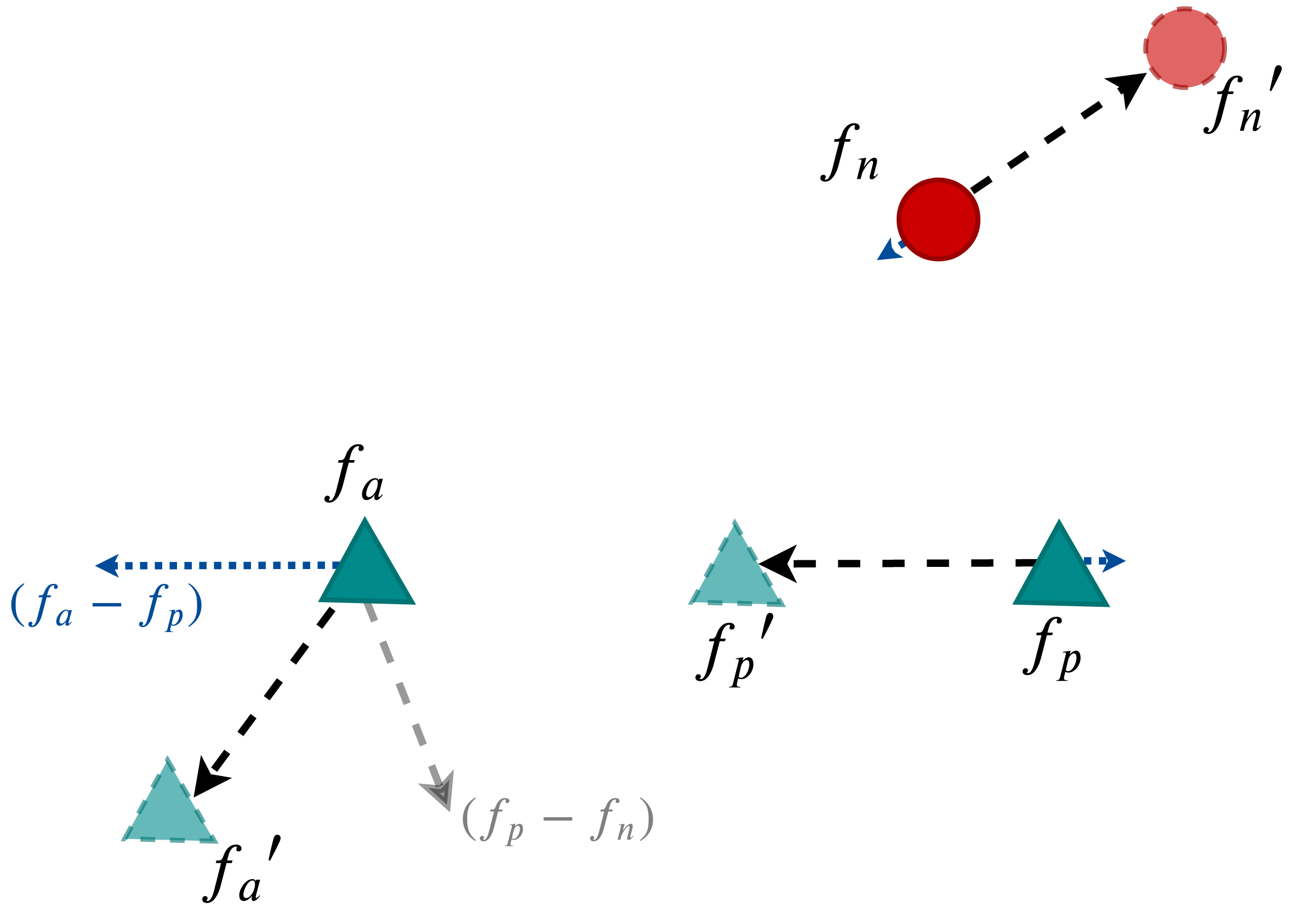}
\caption{Illustration of direction of gradient forces acting on Anchor,Positive and Negative under a triplet formulation}
\label{direction_triplet}
\centering
\end{figure}

The above equations define the vectors used for updating the embeddings as illustrated in Fig \ref{direction_triplet}. As seen in
the figure, for this formulation during gradient descent, the negative sample experiences a force in the direction of
$f_n$ \text{-} $f_a$ which pushes it radially outward with respect to $f_a$ while the positive sample is pulled towards $f_a$ with a force of $f_a \text{-} f_p$. It is interesting to note that even though the Anchor is radially pushed away from the Negative, there is no such radially outward push experienced by the Positive. This is due to the fact that the triplet-based formulations do not explicitly enforce the positive-negative separation.
Another important aspect of the Triplet Loss formulation is the margin $\alpha$. One can think of $\alpha$ as the minimum separation that needs to be achieved between anchor-positive and anchor-negative distances. One can note that if the separation between the pairs is more than $\alpha$ then the loss term goes to zero. Often $\alpha$ is treated as a hyperparameter, which is selected based on different factors such as dataset, network architecture, etc. So in order to create highly separable embedding space, selecting informative triplets of samples is crucial. In this context, informative triplets refer to those triplets in the datasets, whose pairwise distances violate the margin. The process of finding such informative samples to provide better optimization is often referred to as sample mining or sampling. A trivial way of identifying such informative samples is to compute the similarity of feature representations for the entire dataset using the current network model. One can easily create informative triples with such an exhaustive offline process. Although straightforward, this process often becomes computationally infeasible as the size of the dataset increases. Alternatively, sample mining is often done in the mini-batch $B$ by computing similarities only with features of samples present in the mini-batch, thereby restricting the computational cost.

More recently, triplet-based formulations are used with unit-normed feature representations. So the final loss formulation uses cosine similarity instead of Euclidian distance and can be written as:
\begin{equation}
\label{eqn_triplet_cosine}
\begin{split}
\mathcal{L} =  \left[f_a.f_n - f_a.f_p + \alpha) \right]_+
\end{split}
\end{equation}

\subsection{N-Pair Loss}

When analyzing the formulation of Triplet Loss, one can note that during one update, only one negative sample belonging to an arbitrary negative class is chosen. This might be sub-optimal as the update would focus on separating the feature representation of Anchor and Positive from just this Negative sample representation. This might not be ideal as we would like to have feature representations of Anchor and Positive to be well separated from all the other negative classes. Although it is possible that over a large number of mini-batches across multiple epochs, the Anchor and Positive might get well separated from all the other classes, this might not be guaranteed. As a result, most often the Triplet based formulation leads to slower convergence. Additionally, due to the need for mining informative pairs to improve the optimization as mentioned in the last section, most of the randomly sampled triplets are not as useful after the initial phase of training. So, can there be an improved formulation that can help create a better separable feature space? 

One of the potential ways to solve the slow convergence problem is to incorporate multiple negatives into the formulation in Eq.  (\ref{eqn_triplet_loss}). If we consider the set of feature representations  $F_s = \{ f_a^i,f_p^i,f_n^1,f_n^2,.....f_n^j \}$, where $f_a^i$ is a sample belonging to class $i$, $f_p^i$ is a positive sample belonging to the same class, and $f_n^1$,$f_n^2$,....$f_n^j$ represent negative samples belonging to classes 1 through $j$ (excluding class $i$), then given the set of feature representations, Eq (\ref{eqn_triplet_loss}) can be modified to

\begin{equation}
\label{eqn_ntuple_loss}
\begin{split}
\mathcal{L} = \text{log} \left[1 + \sum_{k=1}^{j} (f_a^i.f_n^k - f_a^i.f_p^i) \right] 
\end{split}
\end{equation}
One can note that in this formulation, if a single negative sample is considered, then the loss term in effect reduces down to Eq (\ref{eqn_triplet_cosine})

\begin{figure}[h]
\vspace{2.0em}
\includegraphics[width=12cm]{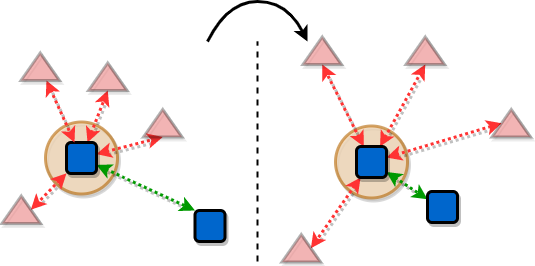}
\caption{N-Pair loss incorporates a larger number of negative samples compared to triplet loss}
\centering
\end{figure}
Although the above formulation solves the problem of slow convergence, it is highly inefficient. If we consider $N$ to be the total number of classes in the dataset, for each update, a set of features of size $(N+1)$ needs to be created, as mentioned above. Given a batch size $B$, this would require $BX(N+1)$ samples to update the parameters of the network in one gradient step.
A better and more efficient method to achieve the same objective was proposed by \cite{sohn2016improved} in which a batch can be constructed more efficiently to reduce the computational cost. Each mini-batch is constructed in such a way that it consists of two samples from each class in the dataset. The new feature set $F_s$ can be written as  
$F_s = \{ (f_a^1,f_p^1),(f_a^2,f_p^2),.....(f_a^n,f_p^n) \}$,
which has a size of $2N$ where $N$ is the number of classes in the dataset. Given the feature set $F_s$, the final optimization objective can be formulated as follows:
\begin{equation}
\label{eqn_ntuple loss}
\begin{split}
\mathcal{L} = \frac{1}{B} \sum^{B}_{i=1} \left\{ \text{log} \left[1 + \sum_{i\neq j} (f_a^i.f_p^j - f_a^i.f_p^i) \right] \right\}
\end{split}
\end{equation}
One can note that only $2N$ embeddings are used to create the necessary feature sets for the optimization which is far more optimal compared to using $BX(N+1)$ embeddings. In practice, often the number of classes in the dataset is larger than the size of the mini-batch. In this scenario, a subset of classes $C$ $<$ $B$  are sampled randomly and used to construct the mini-batch.

\subsection{Multi-Similarity Loss}

As discussed in the previous sections, one of the important aspects of pair-based metric learning is the mining of informative samples. Triplet loss uses the margin $\alpha$ as a measure to mine informative triples, whereas, in the case of N-pair loss, a larger number of negative samples is used to create more informative pairs for optimization. But is there a better way to sample more informative pairs? 
 
 In oder to improve the sampling process, \cite{wang2019multi} proposed analyzing the sample similarities. According to \cite{wang2019multi} sample level similarity can be divided into three different categories. First is self-similarity, which is restricted to the similarity between a pair of samples. For example, if the feature representations  $f_i$ and $f_j$ of two samples in the dataset, belonging to two different classes are highly similar, then this pair of samples becomes highly informative. Here $f_i$ and $f_j$ are called hard negatives with respect to each other. These pairs are said to have high self-similarity and are often identified by using a margin $\alpha$ similar to the one employed by Triplet Loss.

\begin{figure}[h]
\vspace{2.0em}
\includegraphics[width=12cm]{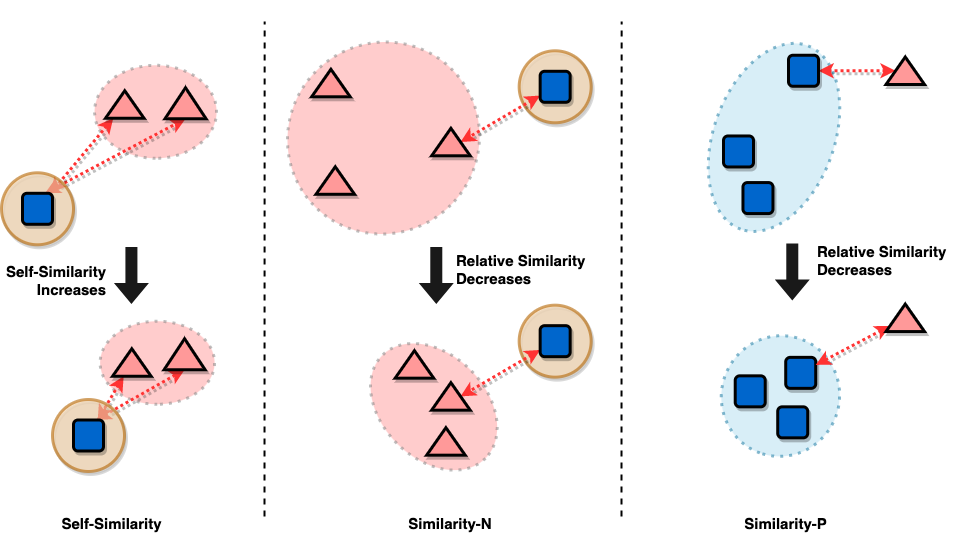}
\caption{MS Loss: Multi-similarity loss computes three different types of similarities (i) Left: Self-similarity (ii) Middle: Negative Relative Similarity (iii) Right: Positive Relative Similarity. Anchor embeddings are enclosed in a brown circle.}
\label{mslossfig}
\centering
\end{figure}
 
 The second type of similarity is called negative relative similarity. This measures the similarity of a pair of samples relative to other negative pairs. If the self-similarity of other negative pairs in the neighborhood of the sample pair is also high then the negative relative similarity of the pair is low. The idea behind negative similarity is to estimate how unique the sample pair is when compared to other negative pairs in the neighborhood (as shown in figure \ref{mslossfig}). So, if there are other negative pairs with equally high self-similarity, then the unique value that this particular sampled pair adds is marginal. This information is captured by the negative relative similarity value. 
 
 Similarly, we can define a positive relative similarity. Let Anchor $f_a$ and Positive $f_p^1$ be the feature representations of two samples belonging to the same class. The relative positive similarity measures whether other positive samples in the neighborhood of $f_p$ also have high self-similarity with $f_a$. In other words, it quantifies how similar a given pair is to other pairs constructed using the same Anchor and other positives of the same class. If they are highly similar, then the current pair under consideration is said to have a low positive relative similarity score.

Based on these three similarity types, \cite{wang2019multi} devises a sampling strategy incorporating all the three sample similarity measures. Consider a mini-batch of size B. Given the feature representation of each sample $f_i$, the similarity between all the pairs of features is obtained as 
\begin{equation}
\label{similarity}
S = \frac{f_i^T f_j}{||f_i||.||f_j||}, \forall i,j \in[1,B] \\
\end{equation}

where the $i^{th}$ row of the similarity matrix, will define the similarity of $i^{th}$ sample with all the other samples in the batch. In order to sample informative pairs, these pairs are first filtered using the positive relative similarity. For this, the positive pair with the lowest similarity value in the $i^{th}$ row is identified. Let us assume $S_{ik}$ corresponds to this value. Now the filtering is done by considering the negative pairs which have a similarity greater than $S_{ik}$. Mathematically,

\begin{equation}
S_{i{n}} > S_{ik} - \epsilon; 
\end{equation}

Similarly, positive pairs are filtered by considering the negative pair with the most similarity to the $i^{th}$ sample. Let this be given by $S_{ij}$
\begin{equation}
{S_{ip}} <  S_{ij} + \epsilon;
\end{equation}
where $S_{ij}$ refers to the highest negative similarity  to the $i^{th}$ sample.

Given the sampled informative pairs, the final formulation of Multi-Similarity loss aims to weight these pairs based on their importance. The final formulation of the loss is given as 

\begin{equation}
\label{eqn_ms_loss}
\begin{split}
\mathcal{L_{MS}} = \frac{1}{B} \sum^{B}_{i=1} \left\{ \frac{1}{\alpha} \, \text{log} \left[1 + \sum_{} e^{-\alpha(S_{ip} - \lambda)}\right] + \right. \\ \left. \frac{1}{\beta} \, \text{log} \left[1 + \sum_{} e^{\beta(S_{in} - \lambda)} \right] \vphantom{\frac{1}{\alpha}} \right\}
\end{split}
\end{equation}

The first $log$ term deals with the cosine similarity scores $S_{ip}$ for the filtered positive samples corresponding to the $i^{th}$ anchor. The second $log$ term analogously deals with that of the filtered negative samples. $\alpha, \beta$ are hyper-parameters used to weight the similarity terms $S_{ip}$ and $S_{in}$ respectively. $\lambda$ acts as the margin term similar to that of Triplet Loss.

\section{Proxy Based Methods}
We have thus far discussed a common approach to deep metric learning which is based on optimizing a sample-to-sample similarity-based objective (such as triplet loss). These objectives are defined in terms of triplets of samples, where a triplet consists of an anchor sample, a positive sample (that is similar to the anchor), and a negative sample (that is dissimilar to the anchor). The objective function is then defined in terms of the distances between the anchor and positive samples, and between the anchor and negative samples. We have referred to these methods as pair based formulations. 

Given a dataset $D$ with $n$ samples, the number of possible triplets with a matching sample and a non-matching sample could be in the order of $O(n^3)$. During each optimization step of stochastic gradient descent, a mini-batch (with $B$ samples) would consist of only a subset of the total number of possible triplets (in the order of $O(B^3)$). Thus, to see all triplets during training the complexity would be in the order of $O(n^3/B^3)$. This impacts the convergence rate since it is highly dependent on how efficiently these triplets are used. This further leads to one of the main challenges in optimizing such a sample-to-sample similarity based objective which is the need to find informative triplets that are effective at driving the learning process. 

To address this challenge, a variety of tricks have been utilized, such as increasing the batch size, using hard or semi-hard triplet mining, utilizing online feature memory banks and other techniques. These approaches aim to improve the quality of the triplets used to optimize the objective function, and can help to improve the performance of the deep metric learning model. 

Proxy based methods have been proposed to overcome this informative pair mining bottleneck of traditional pair-based methods. In particular, these proxy-based methods utilize a set of learnable shared embeddings that act as class representatives during training. Each data-point in the training set is approximated to at least one of the proxies. Unlike pair-based methods, these class representative proxies do not vary with each batch and are shared across samples. Since these proxies learn from all samples in each batch, they eliminate the need for explicitly mining informative pairs.

In the following sections, we will discuss three recent proxy-based loss formulations, namely (i) ProxyNCA \cite{proxynca} and ProxyNCA++  \cite{teh2020proxynca++}(ii) Proxy Anchor Loss \cite{proxyanchor} (iii) ProxyGML Loss \cite{proxyGML}.

\subsection{ProxyNCA and ProxyNCA++}

ProxyNCA is motivated by the seminal work performed by \cite{normalNCA} on Neighbourhood Component Analysis (NCA). Let $p_{ij}$ be the assignment probability or neighborhood probability of $f_i$ to $f_j$, where $f_i, f_j$ are two data points. We can define this probability as:

\begin{equation}
\label{ncapi}
p_{ij} = \frac{ -D_e(f_i, f_j )}{\sum_{k \not\in i} -D_e(f_i, f_k)} 
\end{equation}

where $D_e(f_i, f_k)$ is Euclidean squared distance computed on some learned embeddings. The fundamental purpose of NCA is to increase the probability that points belonging to the same class are close to one another, while simultaneously reducing the probability that points in different classes are near each other. This is formulated as follows:

\begin{equation}
\label{NCAloss}
L_{\text{NCA}} = -\text{log}\left(\frac
{ \sum_{j \not\in C_{i}} e^{-D_e(f_i, f_j)}}
{ \sum_{k \not\in C_{i}} e^{-D_e(f_i, f_k)}} \right)
\vspace{0.5em}
\end{equation}

 The primary challenge in directly optimizing the NCA objective is the computation cost which increases polynomially as the number of samples in the dataset increases. ProxyNCA attempts to address this computation bottleneck of NCA by introducing proxies. Proxies can be interpreted as class representatives or class prototypes. These proxies are implemented as learnable parameters that train along with the feature encoder. During training, instead of computing pair-wise distance which grows quadratically with the batch size and is highly dependent on the quality of pairs, ProxyNCA computes the distance between the learnable set of class proxies and the feature representation of the respective samples in the batch. Based on this and derived from eq. (\ref{NCAloss}), ProxyNCA loss \cite{proxynca} can be formulated as:

\begin{equation}
\label{ProxyNCAloss}
L_{\text{ProxyNCA}} = -\text{log}\left(\frac
{ e^{-D_e(\hat{f_i}, \hat{P_i})}}
{ \sum_{k \not\in C_{i}} e^{-D_e(\hat{f_i}, \hat{P_k})}} \right)
\end{equation}

\begin{figure}[h]
\vspace{2.0em}
\includegraphics[width=12cm]{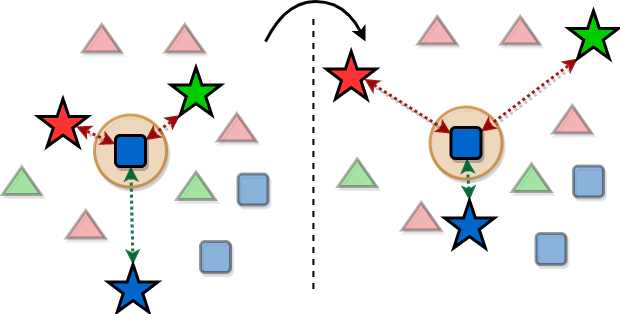}
\caption{Illustration of ProxyNCA Loss: Unlike pair-wise losses that require informative pair mining, ProxyNCA injects learnable proxies denoted by star in the above diagram and pulls them towards the anchored sample embeddings.}
\centering
\end{figure}
where, $P_i$ is the proxy vector corresponding to the data point $f_i$. $\hat{f}$ and $\hat{P}$ denote normalized embeddings. $C_i$ is the set of all data points belonging to the same class as sample $f_i$. The above loss function aims to maximize the distance between non-corresponding proxy and feature pairs and minimize distance between corresponding proxy and feature vectors.

As can be observed in Eq. (\ref{ProxyNCAloss}), ProxyNCA \cite{proxynca} does not directly optimize the proxy assignment probability, rather it optimizes for a suboptimal objective. ProxyNCA++ \cite{teh2020proxynca++} was proposed to overcome this issue. ProxyNCA++ \cite{teh2020proxynca++} computes a proxy assignment probability score $P_i$ defined as:

\begin{equation}
\label{ProxyNCA++}
L_{\text{ProxyNCA++}} = -\text{log}\left(\frac
{ e^{-D_e(\hat{f_i}, \hat{P_i})} * \frac{1}{T}}
{ \sum_{k \in A} e^{-D_e(\hat{f_i}, \hat{P_k})} * \frac{1}{T}} \right)
\end{equation}

where $A$ is the set of all proxies. The subtle difference between ProxyNCA \cite{proxynca} and ProxyNCA++ \cite{teh2020proxynca++} formulation is the explicit computation of proxy assignment probability by performing softmax with respect to distance from all proxies. Another important contribution of ProxyNCA++ \cite{teh2020proxynca++} is the introduction of temperature scaling parameter $T$. As T gets larger, the output of the softmax function that is used to compute the proxy assignment probability tends towards a uniform distribution. With a smaller temperature parameter, the softmax function will lead towards a sharper distribution. Lower temperature increases the distance between probabilities of different classes thereby helping in more refined class boundaries.

\subsection{Proxy Anchor Loss}

Introduction of proxies in ProxyNCA \cite{proxynca} loss overcame the limitations imposed by the need for informative pair mining strategies. But, the objective in ProxyNCA \cite{proxynca} formulation does not explicity optimize for fine-grained sample-to-sample similarity since it indirectly represents it through a proxy-to-sample similarity. Proxy Anchor Loss \cite{proxyanchor} investigated this bottleneck and proposed a novel formulation that takes advantage of both pair-wise and proxy-based formulations and provides a more explicit sample-to-sample similarity based supervision along with proxy based supervision.

Similar to ProxyNCA \cite{proxynca}, Proxy Anchor \cite{proxyanchor} defines proxies as a learnable vector in embedding space. Unlike ProxyNCA \cite{proxynca}, Proxy Anchor  formulation represents each proxy as an anchor and associates it with entire data in a batch. This allows the samples within the batch to interact with one another through their interaction with the proxy anchor. Let, $S(f,p)$ be the similarity between a sample $f$ and a proxy $p$. Similar to Multi-similarity loss formulation (Eq. \ref{eqn_ms_loss}), the proxy anchor \cite{proxyanchor} loss is given as: 

\begin{equation}
\label{proxyanchoreq}
\begin{split}
\mathcal{L} = \frac{1}{|P+|} \sum_{p\in P+} \text{log} \left(1 + \sum_{x \in \mathcal{X}_p^+} e^{-\alpha(S(f,p) - \delta)}\right) + \\
\frac{1}{|P|} \sum_{p\in P} \text{log} \left(1 + \sum_{x \in \mathcal{X}_p^-} e^{\alpha(S(f,p) + \delta)}\right)
\end{split}
\end{equation}

where $\delta > 0$ is a margin, $\alpha > 0$ is a scaling factor, $P$ indicates the set of all proxies, and $P+$ denotes the set of positive proxies of data in the batch. As can be noticed in  Eq. (\ref{proxyanchoreq}), proxy anchor \cite{proxyanchor} utilizes a Log-Sum-Exponent formulation to pull $p$ and its most dissimilar positive example together and to push $p$ and its most similar negative example apart. To understand how Proxy Anchor \cite{proxyanchor} provides a more explicit sample-to-sample similarity based supervision, let us compare the gradients of loss functions for ProxyNCA \cite{proxynca} and Proxy Anchor \cite{proxyanchor}.

\begin{figure}[h]
\vspace{2.0em}
\includegraphics[width=12cm]{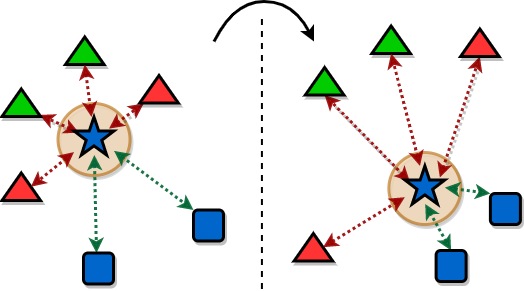}
\caption{Illustration of Proxy Anchor \cite{proxyanchor} Loss: Unlike ProxyNCA, Proxy Anchor \cite{proxyanchor} utilizes a proxy embedding as an anchor and pulls same class sample embeddings closer to it.}
\centering
\end{figure}

The gradient of the loss with respect to $s(f,p)$ for both ProxyNCA and Proxy Anchor \cite{proxyanchor} can be given as:

\begin{equation}
\label{ProxyNCAeqgrad}
\begin{split}
\frac{\partial \mathcal{L_{\text{ProxyNCA}}}}{\partial \bm s(f, p)} = \left\{ 
\begin{array}{l}
-1, \text{if } p = p^+
\\
\frac{e^{s(f,p)}}{\sum\limits_{p^- \in P^-} s(f,p^-)}, \text{ otherwise}
\end{array}
\right\}
\end{split}
\end{equation}

\begin{equation}
\label{proxyanchoreqgrad}
\begin{split}
\frac{\partial \mathcal{L_{\text{ProxyAnchor}}}}{\partial \bm s(f, p)} = \left\{ 
\begin{array}{rl}
\frac{1}{|P^+|} \frac{ - \alpha h^+_p(f)}{1 + \sum\limits_{f' \in F^+_{p}} h^+_p (f')}, \forall f \in F^+_p \\
\frac{1}{|P|} \frac{ - \alpha h^-_p(f)}{1 + \sum\limits_{f' \in F^-_p} h^-_p (f')}, \forall f \in F^-_p
\end{array}
\right\}
\end{split}
\end{equation}

where, $h^{+}_{p}(f)$ = $e^{ - \alpha (s(x,p) - \delta ) }$ and $h^{-}_{p}(f)$ = $e^{\alpha (s(x,p) + \delta ) }$ are positive and negative hardness metrics for embedding vector $f$ given proxy $p$, respectively.

Based on Eq. (\ref{ProxyNCAeqgrad}) and Eq. (\ref{proxyanchoreqgrad}), it can be observed that the gradient of the Proxy-Anchor loss function with respect to the distance measure, $s(f,p)$, is dependent on not just the feature vector $f$, but also on the other samples present in the batch. Then derivation for gradients in Eq. (\ref{ProxyNCAeqgrad}) and Eq. (\ref{proxyanchoreqgrad}) is omitted for brevity, please refer \cite{proxyanchor} for the derivations. This effectively reflects the relative difficulty of the samples within the batch. This is a major advantage of the Proxy-Anchor loss over the Proxy-NCA loss (Eq. \ref{ProxyNCAeqgrad}, which only considers a limited number of proxies when calculating the scale of the gradient for negative examples, and maintains a constant scale for positive examples. On the other hand, the Proxy-Anchor loss determines the scale of the gradient based on the relative difficulty of both positive and negative examples. Additionally, the inclusion of a margin in the loss formulation of the Proxy-Anchor loss leads to improved intra-class compactness and inter-class separability.

\subsection{ProxyGML Loss}

We have discussed two proxy based methods namely, ProxyNCA \cite{proxynca} and Proxy Anchor \cite{proxyanchor}. Despite their differences, one key common aspect of both these methods is the use of a single global proxy per class as a representative prototype. In deep metric learning, we aim to match features from samples belonging to the same class that might be visually very distinct while also distinguishing features from samples belonging to different classes yet visually very similar. Based on this, a single global proxy as a representation for all samples in a class might not be the most optimal method for proxy based optimization.  Proxy-based deep Graph Metric Learning (ProxyGML) \cite{proxyGML} overcame this issue by introducing the notion of multiple trainable proxies for each class that could better represent the local intra-class variations.

Let a dataset $D$ consist of $C$ classes. ProxyGML \cite{proxyGML} assigns $M > 1$ trainable sub-proxies for each class in $C$. So the total number of subproxies in the embedding space is $M$ x $C$. ProxyGML \cite{proxyGML} models the global and local relationships between samples and proxies in a graph like fashion. Here, the directed similarity graph represents the global similarity between all proxies and the samples within the batch. For a sample $f_i$ and proxy $P_j$, the cosine similarity in the global similarity matrix can be defined as:

\begin{equation}
\begin{split}
S_{ij}^p = f_i \cdot P_j
\end{split}
\end{equation}

So the global similarity matrix $S^p$ will have a dimension of ($M$ x $C$) x $B$, since $M$ x $C$ is the total number of proxies and $B$ is the total number of samples in the mini-batch. 

For the $i^{th}$ row in $S^p$ belonging to sample $f_i$, ProxyGML \cite{proxyGML} selects $K$ most similar proxies where $K$ is a hyperparamter. To select $K$ most similar proxies, the method enforces that proxies belonging to the positive class $P^+$ are explicitly selected. The remaining $K-M$ proxies are selected based on their similarity to $f_i$. This results in a new sub-similarity matrix (denoted by $S'$) of dimension $B$ x $K$.

Next, a sub-proxy aggregation on $S'$ is performed by summing the cosine similarities of all proxies that belong to the same class in $C$. This is carried out for each sample in $S'$. Note that, for a sample $f_i$, if no proxy belonging to some class $c$ is present in $S'$ then the class is assigned a similarity of zero. This sub-proxy aggregation strategy results in a final similarity matrix $S$ of dimensions $B$ x $C$.

For small values of $K$, the similarity matrix $S$ can be highly sparse with many zero entries. This leads to an inflated denominator with a traditional softmax operation. Keeping this in mind, a masked softmax operation is used, given by:

\begin{equation}
\begin{split}
P_{ij} = \frac{M_{ij} e^{S_{ij}}}{\sum M_{ij} e^{S_{ij}}}
\end{split}
\end{equation}
where $M$ denotes the zero mask computed as $M_{ij} = 0$ if $S_{ij} = 0$ and $M_{ij} = 1$ if $S_{ij} \neq 0$. Here $P_{ij}$ denotes the softmaxed probability of $j^{th}$ class for the $i^{th}$ sample. Finally, the cross entropy loss is computed over $P_{ij}$ as: 

\begin{figure}[h]
\vspace{2.0em}
\includegraphics[width=12cm]{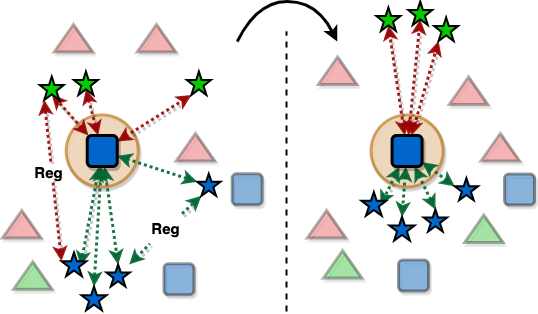}
\caption{Illustration of ProxyGML Loss \cite{proxyGML}: Small stars represent the subproxies. Unlike ProxyNCA \cite{proxynca} and Proxy Anchor \cite{proxyanchor}, ProxyGML \cite{proxyGML} uses multiple proxies per class. Proxies interact with samples and also among themselves in the final loss formulation. Proxies belonging to the same class come closer to one another and the samples belonging to that class, and move away from proxies and samples of other classes.}
\centering
\end{figure}

\begin{equation}
\begin{split}
\mathcal{L_{\text{CE}}}= \frac{1}{B} \sum_{i=0}^{B} \sum_{j=0}^{C} y_{i} \cdot log(P_{ij}) 
\end{split}
\end{equation}

where $y_i$ denotes the ground truth label for the $i^{th}$ sample. Since each class contains multiple sub-proxies which act as class-centers, a third hard constraint is imposed over the similarity between the proxies belonging to the same class. Let $SP$ be the similarity between all $M$ x $C$ number of proxies given as:

\begin{equation}
\label{p2psim}
\begin{split}
SP_{ij} = P_{i} \cdot P_{j}
\end{split}
\end{equation}

A proxy-to-proxy regularization constraint is imposed using Eq. (\ref{p2psim}) by computing the softmax probabilities over $SP$ and then calculating a cross entropy loss as follows:

\begin{equation}
\begin{split}
PP_{ij} = \frac{e^{SP_{ij}}}{\sum e^{SP_{ij}}}
\end{split}
\end{equation}

\begin{equation}
\begin{split}
\mathcal{L_{\text{reg}}}= \frac{1}{M \text{x} C} \sum_{i=0}^{M \text{x} C} \sum_{j=0}^{C} y^p_{i} \cdot log(PP_{ij}) 
\end{split}
\end{equation}

where $PP$ represents the softmax probabilities of proxy-to-proxy similarities and $y^p$ represents ground truth proxy to class label mappings. 

The final loss is computed as:

\begin{equation}
\begin{split}
\mathcal{L} = \mathcal{L_{\text{CE}}} + \lambda \cdot \mathcal{L_{\text{reg}}}
\end{split}
\end{equation}

Here, $\lambda$ denotes the weightage of the regularization term. This approach  demonstrates that end-to-end training by minimizing the final objective $\mathcal{L}$ yields a more discriminative metric space.

\section{Regularizations}
In the previous sections, we explored pair-based and proxy-based loss formulations for creating enhanced feature representations. A majority of these methods have focused on creating new formulations based on sample-sample interaction or sample-proxy interactions. Is there any other information that one can leverage in order to create a more robust embedding space? In this section, we will look at two such sources of information, language and direction, and see how these can be integrated into existing deep metric learning formulations.

\subsection{Language Guidance}
Large Language Models (LLMs) such as BERT \cite{BERT}, ROBERTA \cite{RoBERTa}, etc have been highly successful in modeling and representing content present in the form of natural language. Often these models are trained on a massive corpus of data, whereby they learn to encode proper semantic context and model semantic relationships correctly. On the other hand, training a model with just the deep metric learning objective mentioned in the previous section, using a dataset $X = \{(x_1,y_1),(x_2,y_2)...(x_m,y_m)\}$ in the form of image-label pairs might not encode all the semantic context associated with data points. So, is there a way to leverage the semantic context in the representation of large language models to improve the deep metric learning objective?

\cite{roth2022integrating} proposed a method to incorporate the knowledge from representations of LLMs into a deep metric learning objective. As we know, $X = \{(x_1,y_1),(x_2,y_2)...(x_m,y_m)\}$ represents the set of image-label pairs. Following the discussion in the Multi-Similarity Loss formulation, a similarity matrix $S_I$ of feature representations of data samples are computed using Eq (\ref{similarity}). In order to incorporate the language based regularization factor, prompts are generated from the class labels. Each class label $y_i$ in the batch is used to create a prompt $T_i$ which is represented by 'A photo of $<y_i>$'. Let the set of these prompts corresponding to the batch be $T$. So $T = \{T_1,T_2...T_C\}$ where $C$ is the size of the batch. These prompts are passed through large language model $\Omega$ to get a lower dimensional representation. More formally: 
\begin{equation}
\label{language network}
f_{i}^{lang} = \Omega(T_i;\theta_{\omega}); \forall i \in m
\end{equation}

$\Omega$ here is a pre-trained language model like BERT, Roberta, etc. Once these language representations $f_{i}^{lang}$ are obtained from the prompts, the similarity matrix $S_{L}$ is constructed in a similar fashion to $S_{I}$ using Eq (\ref{similarity}). 

Given the two similarity matrices $S_{I}$ and $S_{L}$, a new distillation loss is proposed to incorporate knowledge from the language modality into the visual modality. The formulation of this loss is as follows:
\begin{equation}
\label{distillation loss}
L_{Lang}= \frac{1}{C} \sum^{C}_{i=1} \left\{ \sigma(S_{I}^i) \, \text{log} \left[\frac{\sigma(S_{I}^i)}{\sigma(S_{L}^i+\gamma_{L})}\right] \right\}
\end{equation}

where $\sigma(S_{I}^i)$ and $\sigma(S_{img}^i)$ represent softmax along each row of the similarity matrix $S_{I}$ and $S_{L}$. Here $\gamma_{L}$ is a language constant set to 1. 
This distillation loss formulation in essence, is a summation of KL divergence between individual rows of $S_{I}$ and $S_{L}$ after converting it into probability distributions using a softmax function.

The final loss formulation is given as a combination of any prior pair-based deep metric learning loss (discussed in the prior section) $L_{DML}$ and $L_{Lang}$, i.e.
\begin{equation}
\label{final loss}
L = L_{DML} + \omega L_{Lang}
\end{equation}
where $\omega$ is a hyperparameter that decides the influence of the language regularization term on the final metric learning objective.

\subsection{Direction Regularization}

The Deep metric learning methods introduced so far, leverage carefully designed sampling strategies and loss formulations that aid in optimizing the generation of a discriminable embedding space. While a lot of work has been done to improve the process of sampling and  loss formulations, relatively less attention has been given to the relative interactions between pairs, and the forces exerted on these pairs that direct their displacement in the embedding space. In order to understand the need for optimal displacement, we revisit the triplet loss and directions in which forces act. One can note that the Negative sample is radially pushed away from the Anchor, during one update. Even though, pushing the negative sample radially away from the anchor intuitively makes sense, it might be a sub-optimal direction due to the presence of other positive samples in that direction. It is also important to note that the Positive sample in the triplet does not have an influence on the direction of force acting on the Negative sample. So is there a more optimal direction in which the forces should act?

\begin{figure}[h]
\vspace{2.0em}
\includegraphics[width=12cm]{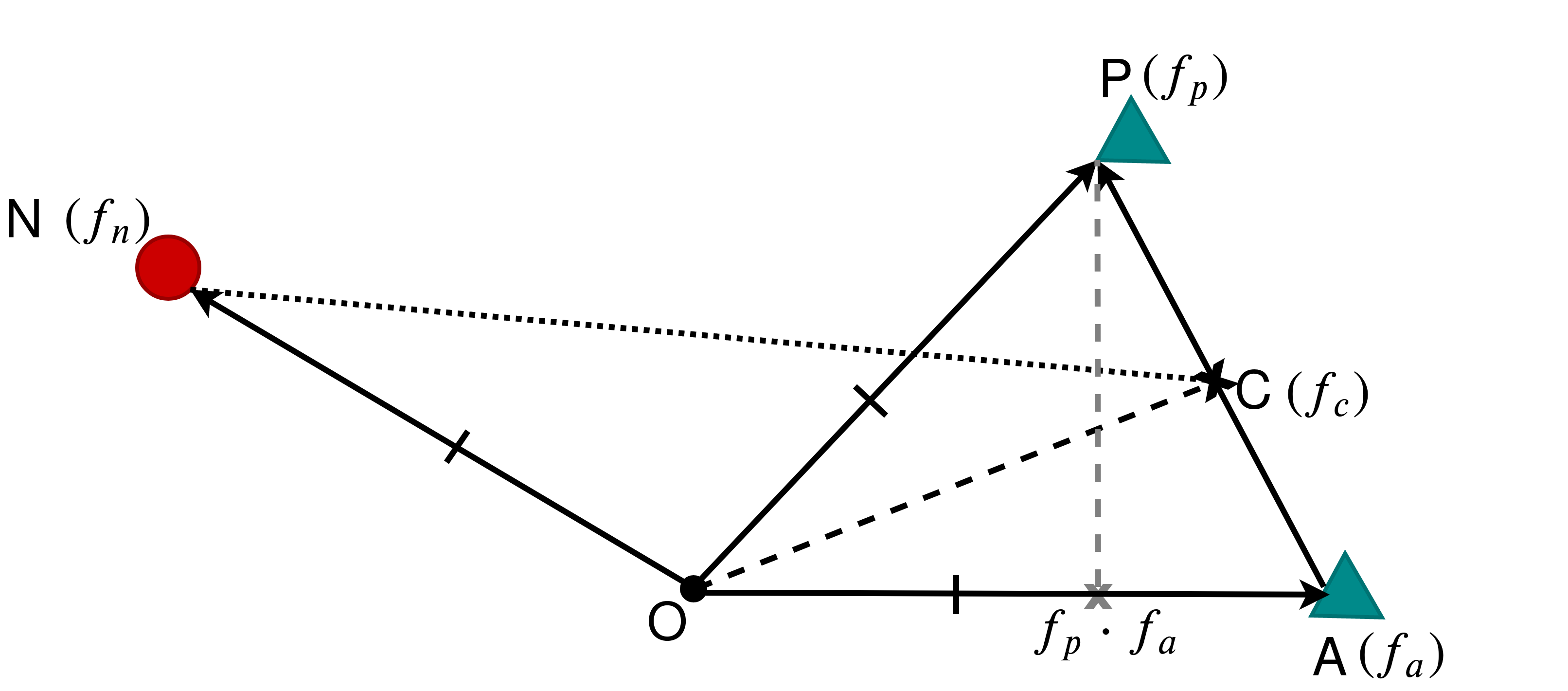}
\caption{Geometric illustration of the layout of the anchor, positive and negative samples. The lines OA, OP and ON represent the unit-normalized embedding vectors for the anchor ($f_a$), positive ($f_p$) and negative ($f_n$) respectively. C is the midpoint of PA and OC represents the average embedding vector $f_c$ (not unit-normalized}
\label{DirectionDMLFig}
\centering
\end{figure}

In figure \ref{direction_triplet}, we see the direction in which the gradient forces act on each sample. In  such a situation, we would additionally
desire to have the negative sample move in the direction orthogonal to the class cluster center of $a$ and $p$ which we approximate as 
$f_c = \frac{f_{a}+f_{p}}{2} $. as shown in Figure \ref{DirectionDMLFig}. Mathematically, we need to enforce the following constraint
\begin{equation}
\label{eqn_nc_dot_pa}
NC \, \bot \, PA \implies \frac{NC}{\|NC\|} \cdot \frac{PA}{\|PA\|} = 0
\end{equation}

Analyzing this in more detail, it turns out that incorporating the cosine similarity between the vectors Anchor-Positive and Anchor-Negative helps achieve this objective. (We are skipping the proof for brevity, please refer to \cite{Mohan_2020_CVPR} for more details). This can be written as 
\begin{equation}
\label{eqn_cos_ap_an}
    Cos(AN, AP) = \frac{1 - \bm f_p \cdot \bm f_a}{\|\bm f_n - \bm f_a\|\|\bm f_p - \bm f_a\|}
\end{equation}
We find that minimizing this cosine similarity term, helps to satisfy the condition in Eq (\ref{eqn_nc_dot_pa}). Since the current metric learning losses lack explicit enforcement of the orthogonality of the negative sample with respect to the anchor-positive pair, when incorporated as a regularization term, it helps create a much more robust embedding space. The following equations provide definitions and the intuition on how easily this regularization term can be adapted into any standard metric learning loss function. The modified triplet loss can be written as:
\begin{equation}
\begin{split}
\label{eqn_directed_triplet_loss_compact}
\mathcal{L}_{apn} = & {\|\bm f_a - \bm f_p\|}^2 - {\|\bm f_a - \bm f_n\|}^2 + \alpha \, \\ 
& - \gamma \, Cos(\bm f_n - \bm f_a, \bm f_p - \bm f_a)
\end{split}
\end{equation}

Similarly, the modified Multi-similarity loss can be written as:
\begin{equation}
\label{eqn_directed_ms_loss}
\begin{split}
\mathcal{L} = \frac{1}{B} \sum^{B}_{i=1} \left\{ \frac{1}{\alpha} \, \text{log} \left[1 + \sum_{} e^{-\alpha(S_{ip} - \lambda)}\right] + \right. \\ \left. \frac{1}{\beta} \, \text{log} \left[1 + \sum_{} e^{\beta(S_{in} - \lambda - \gamma \, Cos(\bm f_n - \bm f_a, \bm f_p - \bm f_a))} \right] \vphantom{\frac{1}{\alpha}} \right\}
\end{split}
\end{equation}
In the case of Multi-Similarity loss, the hardest positive to the anchor is used in the regularization term.

In the case of ProxyNCA \cite{proxynca}, the direction of forces is modified to push negative samples in the direction of their corresponding proxies. It is given by :
\begin{equation}
\label{eqn_directed_proxy_loss}
\mathcal{L} = \sum_{a \subset N} - \text{log} \left( \frac{e^{(- {\|\bm f_a - p(a)\|}^2)}}{\sum_{n} e^{\left[- {\|\bm f_a - p(n)\|}^2 - \gamma \, Cos(p(n) - \bm f_a, p(a) - \bm f_a)\right]}} \right)
\end{equation}
\section{Conclusion}
Deep Metric Learning approaches provide objective functions that help to create highly discriminable embedding spaces. In this chapter, we have provided a brief review of some of the important deep metric learning methods and elaborated on the uniqueness of each approach compared to each other. Currently, owing to the recent advancements in the area of Cross-modal retrieval methods such as CLIP \cite{CLIP}, deep metric learning methods which primarily focused on a single modality are being adapted towards multi-modal representation learning. Recent works such as \cite{wei2020universal} and \cite{wacvnaploss}, show the ability of these methods to create more robust feature representations for multi-modal tasks such as text-image retrieval.
\printbibliography

\end{document}